%% file: main.tex
\documentclass{article}

\usepackage{graphicx}
\usepackage{subfigure}
\usepackage{bbm}
\usepackage{enumitem}
\usepackage{amsmath,amssymb,amsthm}
\usepackage[export]{adjustbox}
\usepackage{changes}
\usepackage{multirow}

\input{defs.tex}
\usepackage[numbers]{natbib}

    \usepackage[preprint]{neurips_2021}

\usepackage[utf8]{inputenc} % allow utf-8 input
\usepackage[T1]{fontenc}    % use 8-bit T1 fonts
\usepackage{hyperref}       % hyperlinks
\usepackage{url}            % simple URL typesetting
\usepackage{booktabs}       % professional-quality tables
\usepackage{amsfonts}       % blackboard math symbols
\usepackage{nicefrac}       % compact symbols for 1/2, etc.
\usepackage{microtype}      % microtypography
\usepackage{xcolor}         % colors

\usepackage{listings}
\usepackage{algorithm}
\usepackage{algpseudocode}

\hypersetup{
 colorlinks=False,
 linkcolor=blue,
 citecolor=blue,
 urlcolor=blue}

\definechangesauthor[color=purple, name=kimin]{KM}

\title{Decision Transformer: Reinforcement \\ Learning via Sequence Modeling}

\author{%
  Lili Chen$^{* , 1}$,
  Kevin Lu$^{* , 1}$,
  \textbf{Aravind Rajeswaran}$^{2}$,
  \textbf{Kimin Lee}$^{1}$,
  \vspace{3pt}
  \\
  \textbf{Aditya Grover}$^{2}$,
  \textbf{Michael Laskin}$^{1}$,
  \textbf{Pieter Abbeel}$^{1}$,
  \textbf{Aravind Srinivas}$^{\dagger, 1}$,
  \textbf{Igor Mordatch}$^{\dagger, 3}$
  \vspace{3pt}
  \\
  $^{*}$equal contribution \vspace{.1em} \hspace{4pt}
  $^{\dagger}$equal advising
  \vspace{3pt}
  \\
  $^{1}$UC Berkeley \vspace{.1em} \hspace{4pt}
  $^{2}$Facebook AI Research \vspace{.1em} \hspace{4pt}
  $^{3}$Google Brain
  \vspace{3pt}
  \\
  \texttt{\{lilichen, kzl\}@berkeley.edu}
}

\begin{document}

\maketitle
\input{sections/abstract}

\clearpage
\tableofcontents
\clearpage

\input{sections/intro}

\input{sections/preliminaries}

\input{sections/method}
\clearpage
\input{sections/experiments}

\input{sections/discussion}

\input{sections/relatedwork}

\input{sections/conclusion}

\input{sections/acknowledgements}

\clearpage  %%%%%%%%%%%%%%%%%%%%%%%%%%%%%%%%%%%%%%%%%%%

\bibliography{reference}
\bibliographystyle{unsrtnat}

\clearpage %%%%%%%%%%%%%%%%%%%%%%%%%%%%%%%%%%%%%%%%%%%

\appendix

% \input{sections/checklist}
% \newpage

% \input{appendix/experimental_details}

\input{appendix/atari_hyperparameters}

\input{appendix/atari_raw}

\end{document}

%% file: defs.tex
\newcommand{\alert}[1]{\textbf{}}

\newcommand{\BEAS}{\begin{eqnarray*}}
\newcommand{\EEAS}{\end{eqnarray*}}
\newcommand{\BEA}{\begin{eqnarray}}
\newcommand{\EEA}{\end{eqnarray}}
\newcommand{\BEQ}{\begin{equation}}
\newcommand{\EEQ}{\end{equation}}
\newcommand{\BIT}{\begin{itemize}}
\newcommand{\EIT}{\end{itemize}}
\newcommand{\BNUM}{\begin{enumerate}}
\newcommand{\ENUM}{\end{enumerate}}
\newcommand{\BEL}[1]{\begin{equation}\label{#1}}
\newcommand{\EEL}{\end{equation}}
\newcommand{\BA}{\begin{array}}
\newcommand{\EA}{\end{array}}

%% file: sections/abstract.tex
\begin{abstract}

We introduce a framework that abstracts Reinforcement Learning (RL) as a sequence modeling problem. This allows us to draw upon the simplicity and scalability of the Transformer architecture, and associated advances in language modeling such as GPT-x and BERT. In particular, we present Decision Transformer, an architecture that casts the problem of RL as conditional sequence modeling. Unlike prior approaches to RL that fit value functions or compute policy gradients, Decision Transformer simply outputs the optimal actions by leveraging a causally masked Transformer. By conditioning an autoregressive model on the desired return (reward), past states, and actions, our Decision Transformer model can generate future actions that achieve the desired return. Despite its simplicity, Decision Transformer matches or exceeds the performance of state-of-the-art model-free offline RL baselines on Atari, OpenAI Gym, and Key-to-Door tasks.

\end{abstract}

\addtocounter{footnote}{-1}
\begin{figure*} [h] \centering
\includegraphics[width=\textwidth]{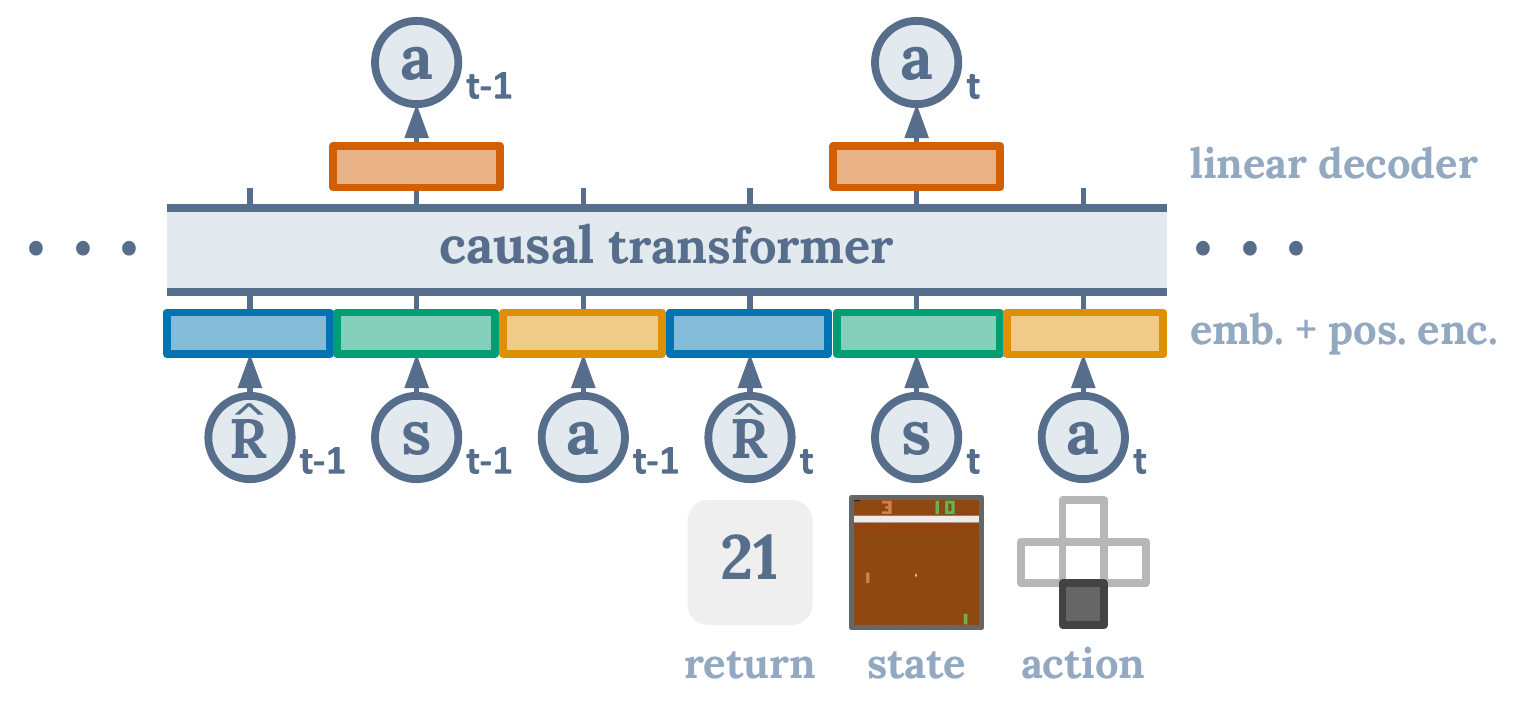}
\caption{
Decision Transformer architecture\protect\footnotemark.
States, actions, and returns are fed into modality-specific linear embeddings and a positional episodic timestep encoding is added.
Tokens are fed into a GPT architecture which predicts actions autoregressively using a causal self-attention mask.
}
\label{fig:main_fig}
\end{figure*}

\footnotetext{
Our code is available at: \texttt{\url{https://sites.google.com/berkeley.edu/decision-transformer}}
}

%% file: sections/intro.tex
\section{Introduction}

Recent work has shown transformers \citep{vaswani2017attention} can model high-dimensional distributions of semantic concepts at scale, including effective zero-shot generalization in language \citep{brown2020gpt3} and out-of-distribution image generation \citep{ramesh2021dalle}.
% This stands in sharp contrast to much work in reinforcement learning (RL), which learns a single policy to model a particular narrow behavior distribution.
Given the diversity of successful applications of such models, we seek to examine their application to sequential decision making problems formalized as reinforcement learning (RL).
In contrast to prior work using transformers as an architectural choice for components within traditional RL algorithms~\cite{parisotto2020stabilizing,zambaldi2018deep}, we seek to study if generative trajectory modeling -- i.e. modeling the joint distribution of the sequence of states, actions, and rewards -- can serve as a \emph{replacement} for conventional RL algorithms.

We consider the following shift in paradigm: instead of training a policy through conventional RL algorithms like temporal difference (TD) learning~\cite{sutton2018reinforcement}, we will train transformer models on collected experience using a sequence modeling objective.
This will allow us to bypass the need for bootstrapping for long term credit assignment -- thereby avoiding one of the ``deadly triad''~\citep{sutton2018reinforcement} known to destabilize RL.
It also avoids the need for discounting future rewards, as typically done in TD learning, which can induce undesirable short-sighted behaviors.
Additionally, we can make use of existing transformer frameworks widely used in language and vision that are easy to scale, utilizing a large body of work studying stable training of transformer models.

In addition to their demonstrated ability to model long sequences, transformers also have other advantages.
Transformers can perform credit assignment directly via self-attention, in contrast to Bellman backups which slowly propagate rewards and are prone to ``distractor'' signals~\citep{hung2019optimizing}.
This can enable transformers to still work effectively in the presence of sparse or distracting rewards.
Finally, empirical evidence suggest that a transformer modeling approach can model a wide distribution of behaviors, enabling better generalization and transfer~\citep{ramesh2021dalle}.

We explore our hypothesis by considering {\em offline RL,} where we will task agents with learning policies from suboptimal data -- producing maximally effective behavior from fixed, limited experience.
This task is traditionally challenging due to error propagation and value overestimation \citep{levine2020offline}.
However, it is a natural task when training with a sequence modeling objective.
By training an autoregressive model on sequences of states, actions, and returns, we reduce policy sampling to autoregressive generative modeling.
We can specify the expertise of the policy -- which ``skill'' to query -- by selecting the desired return tokens, acting as a prompt for generation.

\begin{figure*} [b] \centering
\includegraphics[width=1.00\textwidth]{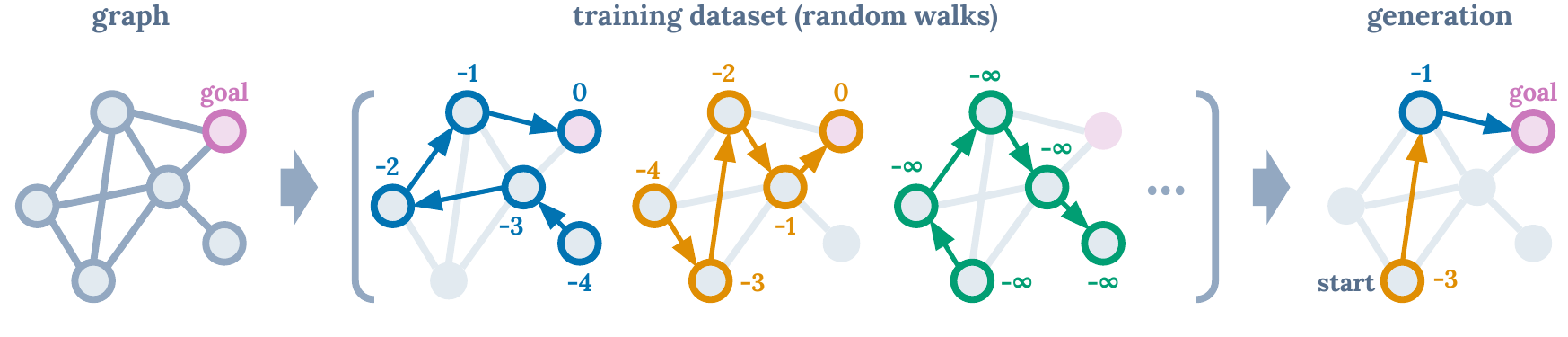}
\caption{
Illustrative example of finding shortest path for a fixed graph (left) posed as reinforcement learning. Training dataset consists of random walk trajectories and their per-node returns-to-go (middle). Conditioned on a starting state and generating largest possible return at each node, Decision Transformer sequences optimal paths.
}
\label{fig:graph_fig}
\end{figure*}

\textbf{Illustrative example.}
To get an intuition for our proposal, consider the task of finding the shortest path on a directed graph, which can be posed as an RL problem. The reward is $0$ when the agent is at the goal node and $-1$ otherwise. We train a GPT \citep{radford2018gpt} model to predict next token in a sequence of returns-to-go (sum of future rewards), states, and actions. Training only on random walk data -- with no expert demonstrations -- we can generate \emph{optimal} trajectories at test time by adding a prior to generate highest possible returns (see more details and empirical results in the Appendix) and subsequently generate the corresponding sequence of actions via conditioning. Thus, by combining the tools of sequence modeling with hindsight return information, we achieve policy improvement without the need for dynamic programming.

Motivated by this observation, we propose Decision Transformer, where we use the GPT architecture to autoregressively model trajectories (shown in Figure \ref{fig:main_fig}).
We study whether sequence modeling can perform policy optimization by evaluating Decision Transformer on offline RL benchmarks in Atari~\citep{bellemare2013arcade}, OpenAI Gym~\citep{brockman2016openai}, and Key-to-Door~\citep{mesnard2020counterfactual} environments.
We show that -- \emph{without using dynamic programming} -- Decision Transformer matches or exceeds the performance of state-of-the-art model-free offline RL algorithms~\citep{agarwal2020optimistic,
kumar2020conservative}.
Furthermore, in tasks where long-term credit assignment is required, Decision Transformer capably outperforms the RL baselines.
With this work, we aim to bridge sequence modeling and transformers with RL, and hope that sequence modeling serves as a strong algorithmic paradigm for RL.

%% file: sections/preliminaries.tex
\section{Preliminaries}

\subsection{Offline reinforcement learning}
We consider learning in a Markov decision process (MDP) described by the tuple ($\mathcal{S}$, $\mathcal{A}$, $P$, $\mathcal{R}$). The MDP tuple consists of states $s \in \mathcal{S}$, actions $a \in \mathcal{A}$, transition dynamics $P(s'|s,a)$, and a reward function $r = \mathcal{R}(s,a)$. 
We use $s_t$, $a_t$, and $r_t=\mathcal{R}(s_t, a_t)$ to denote the state, action, and reward at timestep $t$, respectively. A trajectory is made up of a sequence of states, actions, and rewards: $\tau = (s_0, a_0, r_0, s_1, a_1, r_1, \ldots, s_T, a_T, r_T)$. The return of a trajectory at timestep $t$, $R_t = \sum_{t'=t}^T r_{t'}$, is the sum of future rewards from that timestep.
The goal in reinforcement learning is to learn a policy which maximizes the expected return $\mathbb{E}\Bigl[\sum_{t=1}^T r_t\Bigr]$ in an MDP.
In offline reinforcement learning, instead of obtaining data via environment interactions, we only have access to some fixed limited dataset consisting of trajectory rollouts of arbitrary policies.
This setting is harder as it removes the ability for agents to explore the environment and collect additional feedback.

\subsection{Transformers}
Transformers were proposed by \citet{vaswani2017attention} as an architecture to efficiently model sequential data.
These models consist of stacked self-attention layers with residual connections.
Each self-attention layer receives $n$ embeddings $\{x_i\}_{i=1}^n$ corresponding to unique input tokens, and outputs $n$ embeddings $\{z_i\}_{i=1}^n$, preserving the input dimensions.
The $i$-th token is mapped via linear transformations to a key $k_i$, query $q_i$, and value $v_i$.
The $i$-th output of the self-attention layer is given by weighting the values $v_j$ by the normalized dot product between the query $q_i$ and other keys $k_j$:
\begin{equation}
    z_i = \sum_{j=1}^n \texttt{softmax}(\{\langle q_i, k_{j'} \rangle\}_{j'=1}^n)_j \cdot v_j.
\end{equation}
As we shall see later, this allows the layer to assign ``credit'' by implicitly forming state-return associations via similarity of the query and key vectors (maximizing the dot product).
In this work, we use the GPT architecture \citep{radford2018gpt}, which modifies the transformer architecture with a causal self-attention mask to enable autoregressive generation, replacing the summation/softmax over the $n$ tokens with only the previous tokens in the sequence ($j \in [1, i]$).
We defer the other architecture details to the original papers.

%% file: sections/method.tex
\section{Method} \label{section:method}

In this section, we present Decision Transformer, which models trajectories autoregressively with minimal modification to the transformer architecture, as summarized in Figure \ref{fig:main_fig} and Algorithm \ref{alg:decisiontransformer}.

\textbf{Trajectory representation.}
The key desiderata in our choice of trajectory representation are
that it should enable transformers to learn meaningful patterns and we should be able to conditionally generate actions at test time.
It is nontrivial to model rewards since we would like the model to generate actions based on \emph{future} desired returns, rather than past rewards.
As a result, instead of feeding the rewards directly, we feed the model with the returns-to-go $\widehat{R}_t = \sum_{t'=t}^T r_{t'}$.
This leads to the following trajectory representation which is amenable to autoregressive training and generation:
\begin{equation}
    \tau = \left(\widehat{R}_1, s_1, a_1, \widehat{R}_2, s_2, a_2, \dots, \widehat{R}_T, s_T, a_T\right).
\end{equation}

At test time, we can specify the desired performance (e.g. 1 for success or 0 for failure), as well as the environment starting state, as the conditioning information to initiate generation.
After executing the generated action for the current state, we decrement the target return by the achieved reward and repeat until episode termination.

\textbf{Architecture.}
We feed the last $K$ timesteps into Decision Transformer, for a total of $3K$ tokens (one for each modality: return-to-go, state, or action).
To obtain token embeddings, we learn a linear layer for each modality, which projects raw inputs to the embedding dimension, followed by layer normalization~\citep{ba2016layernorm}. 
For environments with visual inputs, the state is fed into a convolutional encoder instead of a linear layer.
Additionally, an embedding for each timestep is learned and added to each token -- note this is different than the standard positional embedding used by transformers, as one timestep corresponds to three tokens.
The tokens are then processed by a GPT \citep{radford2018gpt} model, which predicts future action tokens via autoregressive modeling.

\textbf{Training.}
We are given a dataset of offline trajectories.
We sample minibatches of sequence length $K$ from the dataset.
The prediction head corresponding to the input token $s_t$ is trained to predict $a_t$ -- either with cross-entropy loss for discrete actions or mean-squared error for continuous actions -- and the losses for each timestep are averaged.
We did not find predicting the states or returns-to-go to improve performance, although
it is easily permissible within our framework (as shown in Section \ref{sec:key_to_door}) and would be an interesting study for future work.

\begin{algorithm}[hb]
\caption{Decision Transformer Pseudocode (for continuous actions)}
\label{alg:decisiontransformer}
\definecolor{codeblue}{rgb}{0.28125,0.46875,0.8125}
\lstset{
    basicstyle=\fontsize{9pt}{9pt}\ttfamily\bfseries,
    commentstyle=\fontsize{9pt}{9pt}\color{codeblue},
    keywordstyle=
}
\begin{lstlisting}[language=python]
# R, s, a, t: returns-to-go, states, actions, or timesteps
# transformer: transformer with causal masking (GPT)
# embed_s, embed_a, embed_R: linear embedding layers
# embed_t: learned episode positional embedding
# pred_a: linear action prediction layer

# main model
def DecisionTransformer(R, s, a, t):
    # compute embeddings for tokens
    pos_embedding = embed_t(t)  # per-timestep (note: not per-token)
    s_embedding = embed_s(s) + pos_embedding
    a_embedding = embed_a(a) + pos_embedding
    R_embedding = embed_R(R) + pos_embedding
    
    # interleave tokens as (R_1, s_1, a_1, ..., R_K, s_K)
    input_embeds = stack(R_embedding, s_embedding, a_embedding)
    
    # use transformer to get hidden states
    hidden_states = transformer(input_embeds=input_embeds)
    
    # select hidden states for action prediction tokens
    a_hidden = unstack(hidden_states).actions
    
    # predict action
    return pred_a(a_hidden)

# training loop
for (R, s, a, t) in dataloader:  # dims: (batch_size, K, dim)
    a_preds = DecisionTransformer(R, s, a, t)
    loss = mean((a_preds - a)**2)  # L2 loss for continuous actions
    optimizer.zero_grad(); loss.backward(); optimizer.step()

# evaluation loop
target_return = 1  # for instance, expert-level return
R, s, a, t, done = [target_return], [env.reset()], [], [1], False
while not done:  # autoregressive generation/sampling
    # sample next action
    action = DecisionTransformer(R, s, a, t)[-1]  # for cts actions
    new_s, r, done, _ = env.step(action)
    
    # append new tokens to sequence
    R = R + [R[-1] - r]  # decrement returns-to-go with reward
    s, a, t = s + [new_s], a + [action], t + [len(R)]
    R, s, a, t = R[-K:], ...  # only keep context length of K
\end{lstlisting}
\end{algorithm}

%% file: sections/experiments.tex
\section{Evaluations on Offline RL Benchmarks}

In this section, we investigate the performance of Decision Transformer relative to dedicated offline RL and imitation learning algorithms. In particular, our primary points of comparison are model-free offline RL algorithms based on TD-learning, since our Decision Transformer architecture is fundamentally model-free in nature as well. Furthermore, TD-learning is the dominant paradigm in RL for sample efficiency, and also features prominently as a sub-routine in many model-based RL algorithms~\cite{Dyna, janner2019mbpo}. We also compare with behavior cloning and variants, since it also involves a likelihood based policy learning formulation similar to ours. The exact algorithms depend on the environment but our motivations are as follows:
\begin{itemize}[leftmargin=*]
    \item \textbf{TD learning}: most of these methods use an action-space constraint or value pessimism, and will be the most faithful comparison to Decision Transformer, representing standard RL methods.
    A state-of-the-art model-free method is Conservative Q-Learning (CQL)~\citep{kumar2020conservative} which serves as our primary comparison. In addition, we also compare against other prior model-free RL algorithms like BEAR~\cite{kumar2019bear} and BRAC~\cite{wu2019brac}.
    
    \item \textbf{Imitation learning}: this regime similarly uses supervised losses for training, rather than Bellman backups.
        We use behavior cloning here, and include a more detailed discussion in Section \ref{sec:pct_bc}.
\end{itemize}

\begin{figure*}[t]
\centering
\includegraphics[width=1.0\linewidth]{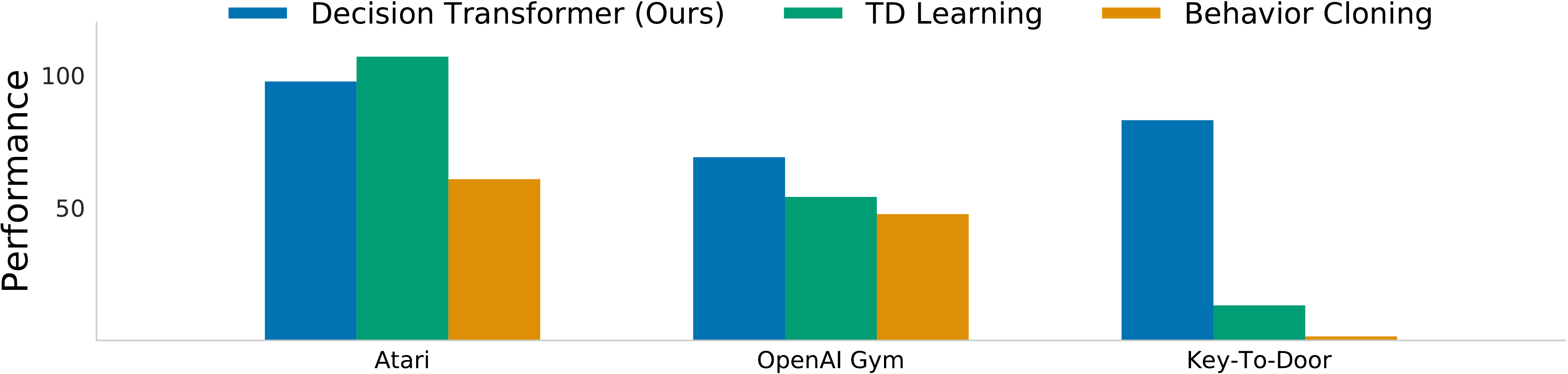}
\caption{
Results comparing Decision Transformer (ours) to TD learning (CQL) and behavior cloning across Atari, OpenAI Gym, and Minigrid.
On a diverse set of tasks, Decision Transformer performs comparably or better than traditional approaches.
Performance is measured by normalized episode return (see text for details).
}
\label{fig:main_results}
\end{figure*}

We evaluate on both discrete (Atari~\citep{bellemare2013arcade}) and continuous (OpenAI Gym~\citep{brockman2016openai}) control tasks.
The former involves high-dimensional observation spaces and requires long-term credit assignment, while the latter requires fine-grained continuous control, representing a diverse set of tasks.
Our main results are summarized in Figure \ref{fig:main_results}, where we show averaged normalized performance for each domain.

\subsection{Atari}

% In this section, we consider Atari games \citep{bellemare2013arcade}, a widely-used RL benchmark.
The Atari benchmark \citep{bellemare2013arcade} is challenging due to its high-dimensional visual inputs and difficulty of credit assignment arising from the delay between actions and resulting rewards.
We evaluate our method on 1\% of all samples in the DQN-replay dataset as per \citet{agarwal2020optimistic}, representing 500 thousand of the 50 million transitions observed by an online DQN agent~\citep{mnih2015human} during training; we report the mean and standard deviation of 3 seeds.
We normalize scores based on a professional gamer, following the protocol of \citet{hafner2020mastering}, where 100 represents the professional gamer score and 0 represents a random policy.

% \begin{table*}[h]
% \centering
% \small
% \begin{tabular}{lrrrrr}
% \toprule
% \multicolumn{1}{c}{\bf Game} & \multicolumn{1}{c}{\bf DT (Ours)} & \multicolumn{1}{c}{\bf CQL} & \multicolumn{1}{c}{\bf QR-DQN} & \multicolumn{1}{c}{\bf REM} & \multicolumn{1}{c}{\bf BC} \\ %  & \multicolumn{1}{c}{\bf BC} \\
% \midrule
% Breakout  & $\bf{267.5} \pm 97.5$ & $211.1$ & $21.1$ & $32.1$ & $138.9 \pm 61.7$ \\ % & $54.6 \pm 2.5$  \\
% Qbert     & $25.1 \pm 18.1$ & $\bf{104.2}$ & $1.7$ & $1.4$  & $17.3 \pm 14.7$ \\ % & $16.7 \pm 8.0$ \\ 
% Pong      & $106.1 \pm 8.1$ & $\bf{111.9}$ & $20.0$ & $39.1$ & $85.2 \pm 20.0$ \\ % &  $2.8 \pm 0.8$  \\
% Seaquest  & $\bf{2.4} \pm 0.7$ & $1.7$ &  $1.4$ &  $1.0$  & $2.1 \pm 0.3$ \\ % &  $0.7 \pm 0.1$\\
% \bottomrule
% \end{tabular}
% \caption{
% Gamer-normalized scores for the 1\% DQN-replay Atari dataset.
% We report the mean and variance across 3 seeds.
% Best mean scores are highlighted in bold.
% Decision Transformer (DT) performs comparably to CQL on 3 out of 4 games, and outperforms other baselines.}
% \label{tbl:atari_main}
% \end{table*}

% updated QR-DQN / REM numbers as per Rishabh Agarwal's email (results from CQL paper were incorrect)
% also using RTG conditioning not based on CQL numbers
\begin{table*}[h]
\centering
\small
\begin{tabular}{lrrrrr}
\toprule
\multicolumn{1}{c}{\bf Game} & \multicolumn{1}{c}{\bf DT (Ours)} & \multicolumn{1}{c}{\bf CQL} & \multicolumn{1}{c}{\bf QR-DQN} & \multicolumn{1}{c}{\bf REM} & \multicolumn{1}{c}{\bf BC} \\ %  & \multicolumn{1}{c}{\bf BC} \\
\midrule
Breakout  & $\bf{267.5} \pm 97.5$ & $211.1$ & $17.1$ & $8.9$ & $138.9 \pm 61.7$ \\ % & $54.6 \pm 2.5$  \\
Qbert     & $15.4 \pm 11.4$ & $\bf{104.2}$ & $0.0$ & $0.0$ & $17.3 \pm 14.7$ \\ % & $16.7 \pm 8.0$ \\ 
Pong      & $106.1 \pm 8.1$ & $\bf{111.9}$ & $18.0$ & $0.5$ & $85.2 \pm 20.0$ \\ % &  $2.8 \pm 0.8$  \\
Seaquest  & $\bf{2.5} \pm 0.4$ & $1.7$ &  $0.4$ & $0.7$  & $2.1 \pm 0.3$ \\ % &  $0.7 \pm 0.1$\\
\bottomrule
\end{tabular}
\caption{
Gamer-normalized scores for the 1\% DQN-replay Atari dataset.
We report the mean and variance across 3 seeds.
Best mean scores are highlighted in bold.
Decision Transformer (DT) performs comparably to CQL on 3 out of 4 games, and outperforms other baselines in most games.}
\label{tbl:atari_main}
\end{table*}

We compare to CQL \citep{kumar2020conservative}, REM \citep{agarwal2020optimistic}, and QR-DQN \citep{dabney2018distributional} on four Atari tasks (Breakout, Qbert, Pong, and Seaquest) that are evaluated in \citet{agarwal2020optimistic}.
We use context lengths of $K=30$ for Decision Transformer (except $K=50$ for Pong).
We also report the performance of behavior cloning (BC), which utilizes the same network architecture and hyperparameters as Decision Transformer but does not have return-to-go conditioning\footnote{We also tried using an MLP with $K=1$ as in prior work, but found this was worse than the transformer.}.
For CQL, REM, and QR-DQN baselines, we report numbers directly from the CQL and REM papers.
We show results in Table \ref{tbl:atari_main}.
Our method is competitive with CQL in 3 out of 4 games and outperforms or matches REM, QR-DQN, and BC on all 4 games.

\subsection{OpenAI Gym} \label{sec:gym_results}

In this section, we consider the continuous control tasks from the D4RL benchmark \citep{fu2020d4rl}.
We also consider a 2D reacher environment that is not part of the benchmark, and generate the datasets using a similar methodology to the D4RL benchmark.
Reacher is a goal-conditioned task and has sparse rewards, so it represents a different setting than the standard locomotion environments (HalfCheetah, Hopper, and Walker).
The different dataset settings are described below.
\begin{enumerate}[leftmargin=*]
    \item Medium: 1 million timesteps generated by a ``medium'' policy that achieves approximately one-third the score of an expert policy.
    \item Medium-Replay: the replay buffer of an agent trained to the performance of a medium policy (approximately 25k-400k timesteps in our environments).
    \item Medium-Expert: 1 million timesteps generated by the medium policy concatenated with 1 million timesteps generated by an expert policy.
\end{enumerate}

We compare to CQL~\citep{kumar2020conservative}, BEAR~\citep{kumar2019bear}, BRAC~\citep{wu2019brac}, and AWR~\citep{peng2019awr}.
CQL represents the state-of-the-art in model-free offline RL, an instantiation of TD learning with value pessimism.
Score are normalized so that 100 represents an expert policy, as per \citet{fu2020d4rl}.
CQL numbers are reported from the original paper; BC numbers are run by us; and the other methods are reported from the D4RL paper.
Our results are shown in Table \ref{tbl:mujoco_results}.
Decision Transformer achieves the highest scores in a majority of the tasks and is competitive with the state of the art in the remaining tasks.

\addtocounter{footnote}{-1}
\begin{table*}[h]
\centering
\small
\begin{tabular}{llrrrrrrr}
\toprule
\multicolumn{1}{c}{\bf Dataset} & \multicolumn{1}{c}{\bf Environment} & \multicolumn{1}{c}{\bf DT (Ours)} & \multicolumn{1}{c}{\bf CQL} & \multicolumn{1}{c}{\bf BEAR} & \multicolumn{1}{c}{\bf BRAC-v} & \multicolumn{1}{c}{\bf AWR} & \multicolumn{1}{c}{\bf BC} \\
\midrule
Medium-Expert & HalfCheetah &  $\bf{86.8} \pm 1.3$ &  $62.4$ &  $53.4$ &  $41.9$ & $52.7$ & $59.9$ \\
Medium-Expert & Hopper      & $107.6 \pm 1.8$ & $\bf{111.0}$ &  $96.3$ &   $0.8$ & $27.1$ & $79.6$ \\
Medium-Expert & Walker      & $\bf{108.1} \pm 0.2$ &  $98.7$ &  $40.1$ &  $81.6$ & $53.8$ & $36.6$ \\
Medium-Expert & Reacher     &  $\bf{89.1 \pm 1.3}$ &  $30.6$ &       - &       - &      - & $73.3$ \\
\midrule
Medium        & HalfCheetah &  $42.6 \pm 0.1$ &  $44.4$ &  $41.7$ &  $\bf{46.3}$ & $37.4$ & $43.1$ \\
Medium        & Hopper      &  $\bf{67.6} \pm 1.0$ &  $58.0$ &  $52.1$ &  $31.1$ & $35.9$ & $63.9$ \\
Medium        & Walker      &   $74.0 \pm 1.4$ &  $79.2$ &  $59.1$ & $\bf{81.1}$ & $17.4$ & $77.3$ \\
Medium        & Reacher     &  $\bf{51.2 \pm 3.4}$ &     $26.0$ &       - &    - &      - & $\bf{48.9}$  \\
\midrule
Medium-Replay & HalfCheetah &  $36.6 \pm 0.8$ &  $46.2$ &  $38.6$ &  $\bf{47.7}$ & $40.3$ & $4.3$ \\
Medium-Replay & Hopper      &  $\bf{82.7 \pm 7.0}$ &  $48.6$ &  $33.7$ &   $0.6$ & $28.4$ & $27.6$ \\
Medium-Replay & Walker      &  $\bf{66.6 \pm 3.0}$ &  $26.7$ &  $19.2$ &   $0.9$ & $15.5$ & $36.9$ \\
Medium-Replay & Reacher     &  $\bf{18.0 \pm 2.4}$ & $\bf{19.0}$ &       - &   - &      - & $5.4$ \\
\midrule
\multicolumn{2}{c}{\bf Average (Without Reacher)} &  $\bf{74.7}$      & $63.9$ & $48.2$ & $36.9$ & $34.3$ & $46.4$ \\
\multicolumn{2}{c}{\bf Average (All Settings)}    &  $\bf{69.2}$      & $54.2$ &    -   &    -   & - & $47.7$ \\
\bottomrule
\end{tabular}
\caption{
Results for D4RL datasets\protect\footnotemark.
We report the mean and variance for three seeds.
Decision Transformer (DT) outperforms conventional RL algorithms on almost all tasks.}
\label{tbl:mujoco_results}
\end{table*}
\footnotetext{Given that CQL is generally the strongest TD learning method, for Reacher we only run the CQL baseline.}

%% file: sections/discussion.tex
\section{Discussion}

\subsection{Does Decision Transformer perform behavior cloning on a subset of the data?}
\label{sec:pct_bc}

In this section, we seek to gain insight into whether Decision Transformer can be thought of as performing imitation learning on a subset of the data with a certain return.
To investigate this, we propose a new method, Percentile Behavior Cloning (\%BC), where we run behavior cloning on only the top $X\%$ of timesteps in the dataset, ordered by episode returns.
The percentile $X\%$ interpolates between standard BC ($X=100\%$) that trains on the entire dataset and only cloning the best observed trajectory ($X\to0\%$), trading off between better generalization by training on more data with training a specialized model that focuses on a desirable subset of the data.

We show full results comparing \%BC to Decision Transformer and CQL in Table \ref{tbl:pct_bc}, sweeping over $X \in [10\%, 25\%, 40\%, 100\%]$.
Note that the only way to choose the optimal subset for cloning is to evaluate using rollouts from the environment, so \%BC is not a realistic approach; rather, it serves to provide insight into the behavior of Decision Transformer.
When data is plentiful -- as in the D4RL regime -- we find \%BC can match or beat other offline RL methods.
On most environments, Decision Transformer is competitive with the performance of the best \%BC, indicating it can hone in on a particular subset after training on the entire dataset distribution.

\begin{table*}[h]
\centering
\small
\begin{tabular}{llrrrrrr}
\toprule
\multicolumn{1}{c}{\bf Dataset} & \multicolumn{1}{c}{\bf Environment} & \multicolumn{1}{c}{\bf DT (Ours)} & \multicolumn{1}{c}{\bf 10\%BC} & \multicolumn{1}{c}{\bf 25\%BC} & \multicolumn{1}{c}{\bf 40\%BC} & \multicolumn{1}{c}{\bf 100\%BC} & \multicolumn{1}{c}{\bf CQL} \\
\midrule
Medium        & HalfCheetah &  $42.6 \pm 0.1$  & $42.9$ & $43.0$ & $43.1$ & $43.1$ & $\bf 44.4$ \\
Medium        & Hopper      &  $\bf 67.6 \pm 1.0$  & $65.9$ & $65.2$ & $65.3$ & $63.9$ & $58.0$ \\
Medium        & Walker      &  $74.0 \pm 1.4$  & $78.8$ & $\bf 80.9$ & $78.8$ & $77.3$ & $79.2$ \\
Medium        & Reacher     &  $51.2 \pm 3.4$  & $51.0$ & $48.9$ & $58.2$ & $\bf 58.4$ & $26.0$ \\
\midrule
Medium-Replay & HalfCheetah &  $36.6 \pm 0.8$  & $40.8$ & $40.9$ & $41.1$ &  $4.3$ & $\bf 46.2$ \\
Medium-Replay & Hopper      &  $\bf 82.7 \pm 7.0$  & $70.6$ & $58.6$ & $31.0$ & $27.6$ & $48.6$ \\
Medium-Replay & Walker      &  $66.6 \pm 3.0$  & $\bf 70.4$ & $67.8$ & $67.2$ & $36.9$ & $26.7$ \\
Medium-Replay & Reacher     &  $18.0 \pm 2.4$  & $\bf 33.1$ & $16.2$ & $10.7$ &  $5.4$ & $19.0$ \\
\midrule
\multicolumn{2}{c}{\bf Average} &  $56.1$ & $\bf{56.7}$ & $52.7$ & $49.4$ & $39.5$ & $43.5$ \\
\bottomrule
\end{tabular}
\caption{
Comparison between Decision Transformer (DT) and Percentile Behavior Cloning (\%BC).}
\label{tbl:pct_bc}
\end{table*}

In contrast, when we study low data regimes -- such as Atari, where we use 1\% of a replay buffer as the dataset -- \%BC is weak (shown in Table \ref{tbl:atari_percentile_bc}).
This suggests that in scenarios with relatively low amounts of data, Decision Transformer can outperform \%BC by using all trajectories in the dataset to improve generalization, even if those trajectories are dissimilar from the return conditioning target.
Our results indicate that Decision Transformer can be more effective than simply performing imitation learning on a subset of the dataset.
On the tasks we considered, Decision Transformer either outperforms or is competitive to \%BC, without the confound of having to select the optimal subset.

\begin{table*}[h]
\centering
\small
\begin{tabular}{lrrrrrr}
\toprule
\multicolumn{1}{c}{\bf Game} & \multicolumn{1}{c}{\bf DT (Ours)} & \multicolumn{1}{c}{\bf 10\%BC} & \multicolumn{1}{c}{\bf 25\%BC} & \multicolumn{1}{c}{\bf 40\%BC} & \multicolumn{1}{c}{\bf 100\%BC} \\
  \midrule
Breakout  & $\bf{267.5} \pm 97.5$ & $28.5 \pm 8.2$ & $73.5 \pm 6.4$ & $108.2 \pm 67.5$ & $138.9 \pm 61.7$ \\
Qbert    & $15.4 \pm 11.4$ & $6.6 \pm 1.7$ & $16.0 \pm 13.8$ & $11.8 \pm 5.8$ & $\bf{17.3} \pm 14.7$ \\
Pong      & $\bf{106.1} \pm 8.1$ & $2.5 \pm 0.2$ & $13.3 \pm 2.7$ & $72.7 \pm 13.3$ & $85.2 \pm 20.0$ \\
Seaquest  & $\bf{2.5} \pm 0.4$ & $1.1 \pm 0.2$ & $1.1 \pm 0.2$ & $1.6 \pm 0.4$ & $2.1 \pm 0.3$ \\
\bottomrule
\end{tabular}
\caption{\%BC scores for Atari. We report the mean and variance across 3 seeds.
Decision Transformer (DT) outperforms all versions of \%BC in most games.}
\label{tbl:atari_percentile_bc}
\end{table*}

\subsection{How well does Decision Transformer model the distribution of returns?}

\begin{figure*}[h]
\centering
\includegraphics[width=1\linewidth]{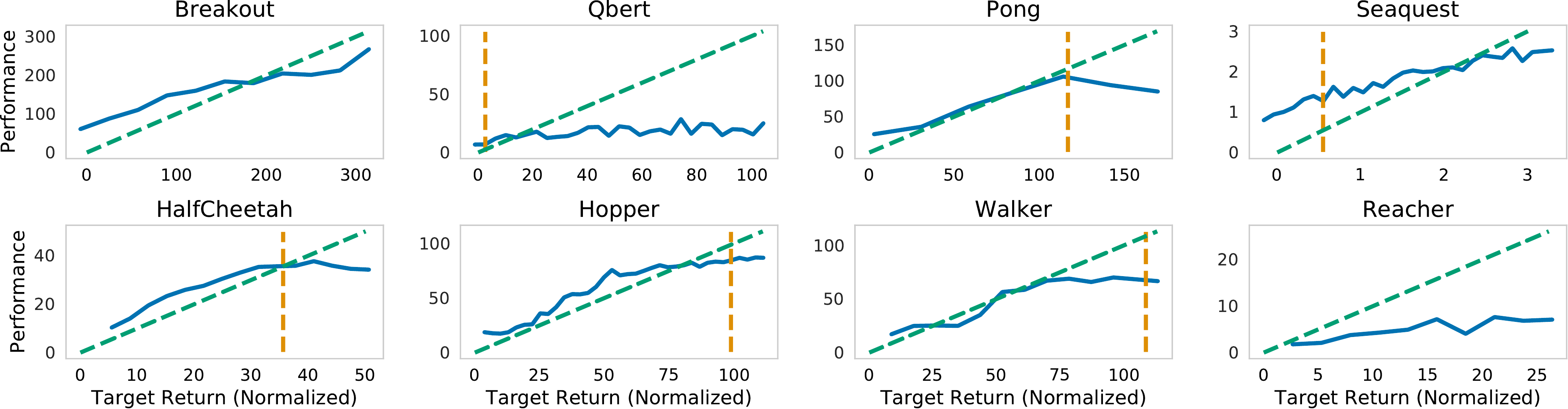} \\
\vspace{.5em}
\includegraphics[width=0.7\linewidth]{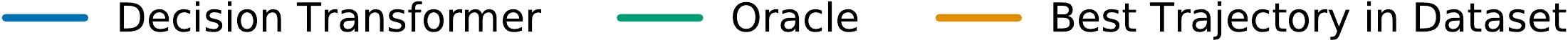}
\caption{
Sampled (evaluation) returns accumulated by Decision Transformer when conditioned on the specified target (desired) returns.
\textbf{Top:} Atari.
\textbf{Bottom:} D4RL medium-replay datasets.
}
\label{fig:atari_return}
\end{figure*}

We evaluate the ability of Decision Transformer to understand return-to-go tokens by varying the desired target return over a wide range -- evaluating the multi-task distribution modeling capability of transformers.
Figure \ref{fig:atari_return} shows the average sampled return accumulated by the agent over the course of the evaluation episode for varying values of target return.
On every task, the desired target returns and the true observed returns are highly correlated. On some tasks like Pong, HalfCheetah and Walker, Decision Transformer generates trajectories that almost perfectly match the desired returns (as indicated by the overlap with the oracle line).
Furthermore, on some Atari tasks like Seaquest, we can prompt the Decision Transformer with higher returns than the maximum episode return available in the dataset, demonstrating that Decision Transformer is sometimes capable of extrapolation.

\subsection{What is the benefit of using a longer context length?}
\label{sec:context_atari}

To assess the importance of access to previous states, actions, and returns, we ablate on the context length $K$.
This is interesting since it is generally considered that the previous state (i.e. $K=1$) is enough for reinforcement learning algorithms when frame stacking is used, as we do.
Table \ref{tbl:atari_history} shows that performance of Decision Transformer is significantly worse when $K=1$, indicating that past information is useful for Atari games.
One hypothesis is that when we are representing a distribution of policies -- like with sequence modeling -- the context allows the transformer to identify which policy generated the actions, enabling better learning and/or improving the training dynamics.

\begin{table*}[h]
\centering
\small
\begin{tabular}{lrrr}
\toprule
\multicolumn{1}{c}{\bf Game} & \multicolumn{1}{c}{\bf DT (Ours)} & \multicolumn{1}{c}{\bf DT with no context ($K=1$)} &  \\
  \midrule
Breakout  & $\bf{267.5} \pm 97.5$ & $73.9 \pm 10$  \\
Qbert    & $\bf{15.1} \pm 11.4$ & $13.6 \pm 11.3$ \\
Pong      & $\bf{106.1} \pm 8.1$ & $2.5 \pm 0.2$ \\
Seaquest  & $\bf{2.5} \pm 0.4$ & $0.6 \pm 0.1$ \\
\bottomrule
\end{tabular}
\caption{Ablation on context length.
Decision Transformer (DT) performs better when using a longer context length ($K=50$ for Pong, $K=30$ for others).}
\label{tbl:atari_history}
\end{table*}

\subsection{Does Decision Transformer perform effective long-term credit assignment?}
\label{sec:key_to_door}

To evaluate long-term credit assignment capabilities of our model, we consider a variant of the Key-to-Door environment proposed in \citet{mesnard2020counterfactual}.
This is a grid-based environment with a sequence of three phases: (1) in the first phase, the agent is placed in a room with a key; (2) then, the agent is placed in an empty room; (3) and finally, the agent is placed in a room with a door.
The agent receives a binary reward when reaching the door in the third phase, but \textbf{only} if it picked up the key in the first phase.
This problem is difficult for credit assignment because credit must be propagated from the beginning to the end of the episode, skipping over actions taken in the middle.

We train on datasets of trajectories generated by applying random actions and report success rates in Table \ref{tbl:key_to_door}.
Furthermore, for the Key-to-Door environment we use the entire episode length as the context, rather than having a fixed content window as in the other environments.
Methods that use highsight return information: our Decision Transformer model and \%BC (trained only on successful episodes) are able to learn effective policies -- producing near-optimal paths, despite only training on random walks.
TD learning (CQL) cannot effectively propagate Q-values over the long horizons involved and gets poor performance.
\clearpage

\begin{table*}[h]
\centering
\small
\begin{tabular}{lrrrrr}
\toprule
\multicolumn{1}{c}{\bf Dataset} & \multicolumn{1}{c}{\bf DT (Ours)} & \multicolumn{1}{c}{\bf CQL} & \multicolumn{1}{c}{\bf BC} & \multicolumn{1}{c}{\bf \%BC} & \multicolumn{1}{c}{\bf Random}\\
\midrule
1K Random Trajectories & $\bf{71.8}\%$ & $13.1\%$ & $1.4\%$ & $69.9\%$ & $3.1\%$      \\
10K Random Trajectories & $94.6\%$ & $13.3\%$ & $1.6\%$ & $\bf{95.1}\%$ & $3.1\%$      \\
\bottomrule
\end{tabular}
\caption{
Success rate for Key-to-Door environment.
Methods using hindsight (Decision Transformer, \%BC) can learn successful policies, while TD learning struggles to perform credit assignment.
}
\label{tbl:key_to_door}
\end{table*}

\subsection{Can transformers be accurate critics in sparse reward settings?}

\begin{figure*}[h]
\centering
\includegraphics[width=0.40\textwidth]{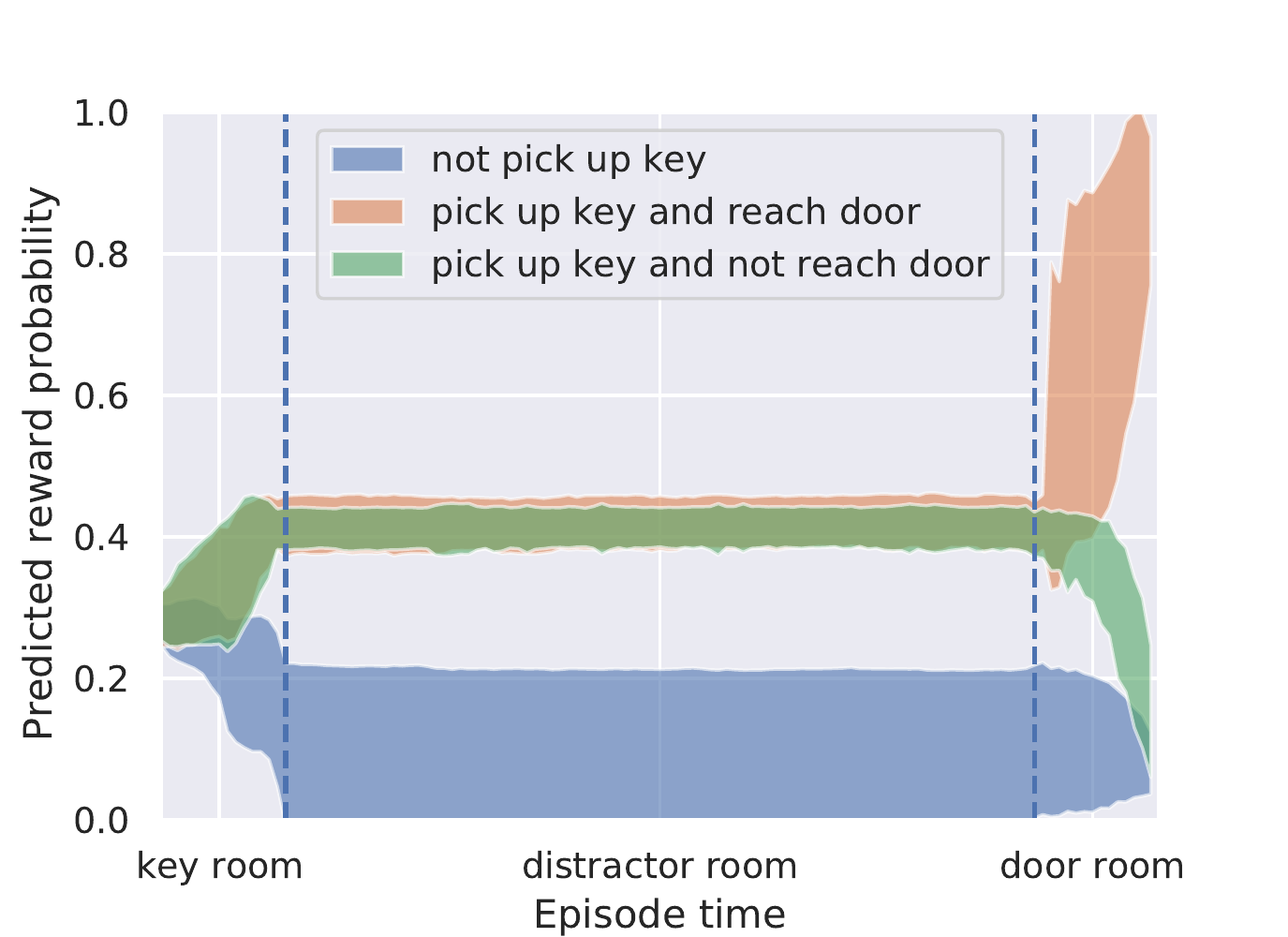}
\includegraphics[width=0.40\textwidth]{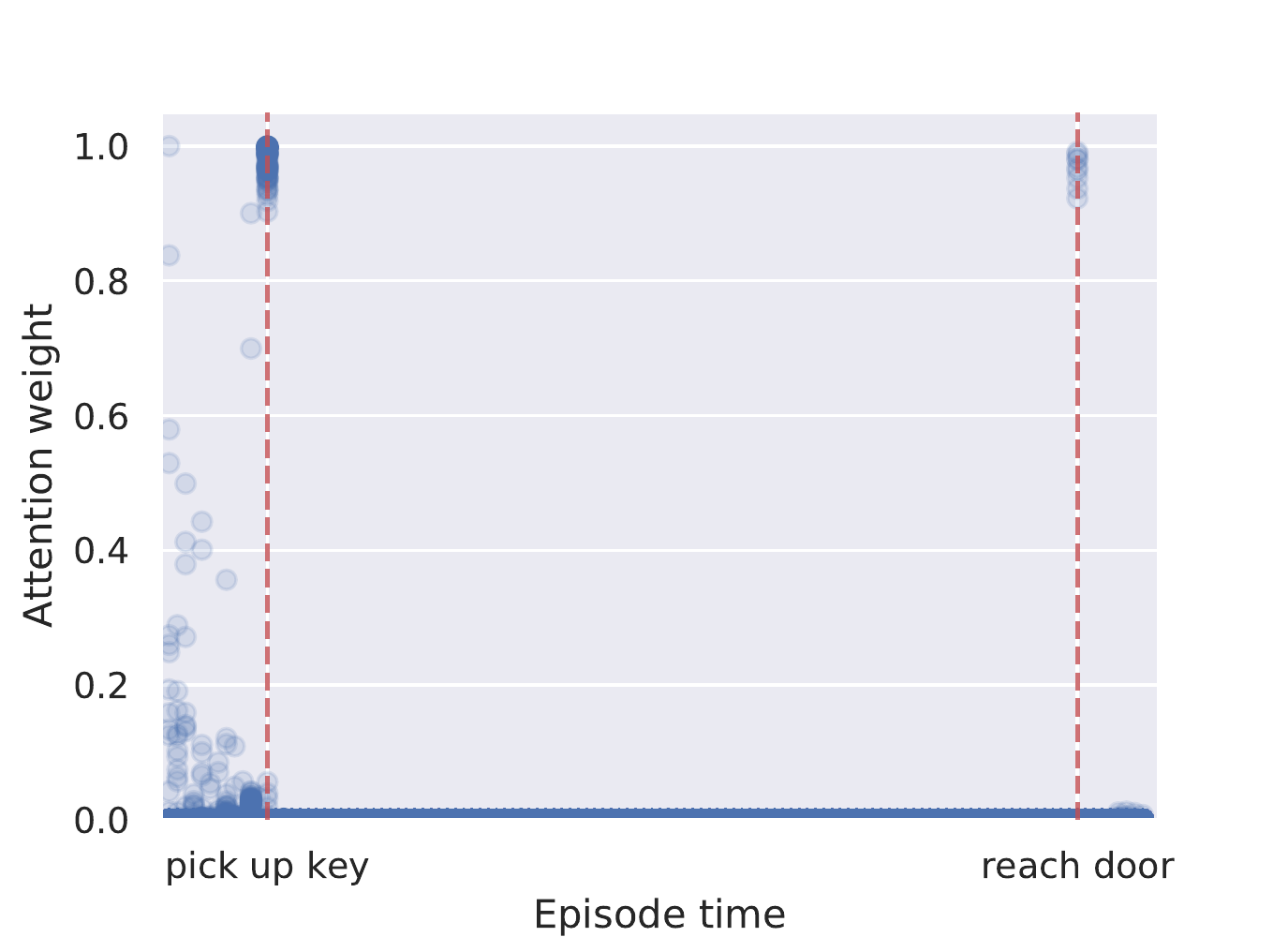}
\caption{
\textbf{Left:} Averages of running return probabilities predicted by the transformer model for three types of episode outcomes.
\textbf{Right:} Transformer attention weights from all timesteps superimposed for a particular successful episode.
The model attends to steps near pivotal events in the episode, such as picking up the key and reaching the door.}
\label{fig:key_critic}
\end{figure*}

In previous sections, we established that decision transformer can produce effective policies (actors).
We now evaluate whether transformer models can also be effective critics.
We modify Decision Transformer to output return tokens in addition to action tokens on the Key-to-Door environment.
Additionally, the first return token is not given, but it is predicted instead (i.e. the model learns the initial distribution $p(\hat{R}_1)$), similar to standard autoregressive generative models.
We find that the transformer continuously updates reward probability based on events during the episode, shown in Figure \ref{fig:key_critic} (Left).
Furthermore, we find the transformer attends to critical events in the episode (picking up the key or reaching the door), shown in Figure \ref{fig:key_critic} (Right), indicating formation of state-reward associations as discussed in \citet{raposo2021synthetic} and enabling accurate value prediction.

\subsection{Does Decision Transformer perform well in sparse reward settings?}

A known weakness of TD learning algorithms is that they require densely populated rewards in order to perform well, which can be unrealistic and/or expensive.
In contrast, Decision Transformer can improve robustness in these settings since it makes minimal assumptions on the density of the reward.
To evaluate this, we consider a delayed return version of the D4RL benchmarks where the agent does not receive any rewards along the trajectory, and instead receives the cumulative reward of the trajectory in the final timestep.
Our results for delayed returns are shown in Table \ref{tbl:mujoco_delayed_results}.
Delayed returns minimally affect Decision Transformer; and due to the nature of the training process, while imitation learning methods are reward agnostic.
While TD learning collapses, Decision Transformer and \%BC still perform well, indicating that Decision Transformer can be more robust to delayed rewards.

\begin{table*}[h]
\centering
\small
\begin{tabular}{llrr|rr|rr}
\toprule
& & \multicolumn{2}{c|}{Delayed (Sparse)} & \multicolumn{2}{c|}{Agnostic} & \multicolumn{2}{c}{Original (Dense)} \\
\multicolumn{1}{c}{\bf Dataset} & \multicolumn{1}{c}{\bf Environment} & \multicolumn{1}{c}{\bf DT (Ours)} & \multicolumn{1}{c|}{\bf CQL} & \multicolumn{1}{c}{\bf BC} & \multicolumn{1}{c|}{\bf \%BC} & \multicolumn{1}{c}{\bf DT (Ours)} & \multicolumn{1}{c}{\bf CQL} \\
\midrule
Medium-Expert & Hopper      & $\bf{107.3 \pm 3.5}$ & $9.0$ & $59.9$ & $102.6$     & $107.6$ & $111.0$ \\
Medium        & Hopper      &  $60.7 \pm 4.5$      & $5.2$ & $63.9$ & $\bf{65.9}$ &  $67.6$ &  $58.0$ \\
Medium-Replay & Hopper      &  $\bf{78.5 \pm 3.7}$ & $2.0$ & $27.6$ & $70.6$      &  $82.7$ &  $48.6$ \\
\bottomrule
\end{tabular}
\caption{
Results for D4RL datasets with delayed (sparse) reward.
Decision Transformer (DT) and imitation learning are minimally affected by the removal of dense rewards, while CQL fails.}
\label{tbl:mujoco_delayed_results}
\end{table*}

\subsection{Why does Decision Transformer avoid the need for value pessimism or behavior regularization?}

One key difference between Decision Transformer and prior offline RL algorithms is that we do not require policy regularization or conservatism to achieve good performance.
Our conjecture is that TD-learning based algorithms learn an approximate value function and improve the policy by optimizing this value function.
This act of optimizing a learned function can exacerbate and exploit any inaccuracies in the value function approximation, causing failures in policy improvement.
Since Decision Transformer does not require explicit optimization using learned functions as objectives, it avoids the need for regularization or conservatism.

\subsection{How can Decision Transformer benefit online RL regimes?}

Offline RL and the ability to model behaviors has the potential to enable sample-efficient online RL for downstream tasks.
Works studying the transition from offline to online generally find that likelihood-based approaches, like our sequence modeling objective, are more successful~\citep{gao2018normalizedpg, nair2020awac}.
As a result, although we studied offline RL in this work, we believe Decision Transformer can meaningfully improve online RL methods by serving as a strong model for behavior generation. For instance, Decision Transformer can serve as a powerful ``memorization engine'' and in conjunction with powerful exploration algorithms like Go-Explore~\cite{ecoffet2019goexplore}, has the potential to simultaneously model and generative a diverse set of behaviors.

%% file: sections/relatedwork.tex
\section{Related Work}

\subsection{Offline reinforcement learning}

To mitigate the impact of distribution shift in offline RL, prior algorithms either (a) constrain the policy action space~\citep{fujimoto2019off, kumar2019stabilizing, siegel2020keep} or (b) incorporate value pessimism~\citep{fujimoto2019off,kumar2020conservative}, or (c) incorporate pessimism into learned dynamics models~\cite{kidambi2020morel, yu2020mopo}.
Since we do not use Decision Transformers to explicitly learn the dynamics model, we primarily compare against model-free algorithms in our work; in particular, adding a dynamics model tends to improve the performance of model-free algorithms.
Another line of work explores learning wide behavior distribution from an offline dataset by learning a task-agnostic set of skills, either with likelihood-based approaches \citep{ajay2020opal, campos2020edl,pertsch2020spirl,singh2021parrot} or by maximizing mutual information \citep{eysenbach2018diversity, lu2020lisp, sharmaGLKH20dads}.
Our work is similar to the likelihood-based approaches, which do not use iterative Bellman updates -- although we use a simpler sequence modeling objective instead of a variational method, and use rewards for conditional generation of behaviors.

\subsection{Supervised learning in reinforcement learning settings}

Some prior methods for reinforcement learning bear more resemblance to static supervised learning, such as Q-learning \citep{watkins1989qlearning, mnih2013dqn}, which still uses iterative backups, or likelihood-based methods such as behavior cloning, which do not (discussed in previous section).
Recent work \citep{srivastava2019udrl, kumar2019rcp, ogma2019blog} studies ``upside-down'' reinforcement learning (UDRL), which are similar to our method in seeking to model behaviors with a supervised loss conditioned on the target return.
A key difference in our work is the shift of motivation to sequence modeling rather than supervised learning: while the practical methods differ primarily in the context length and architecture, sequence modeling enables behavior modeling even without access to the reward, in a similar style to language \citep{radford2018gpt} or images \citep{chen2020igpt}, and is known to scale well \citep{brown2020gpt3}.
The method proposed by \citet{kumar2019rcp} is most similar to our method with $K=1$, which we find sequence modeling/long contexts to outperform (see Section \ref{sec:context_atari}).
\citet{ghosh2019gcsl} extends prior UDRL methods to use state goal conditioning, rather than rewards, and \citet{paster2020glamor} further use an LSTM with state goal conditioning for goal-conditoned online RL settings.

Concurrent to our work, \citet{janner2021tto} propose Trajectory Transformer, which is similar to Decision Transformer but additionally uses state and return prediction, as well as discretization, which incorporates model-based components.
We believe that their experiments, in addition to our results, highlight the potential for sequence modeling to be a generally applicable idea for reinforcement learning.

\subsection{Credit assignment}

Many works have studied better credit assignment via state-association, learning an architecture which decomposes the reward function such that certain ``important'' states comprise most of the credit \citep{ferret2019self, harutyunyan2019hindsight,  mesnard2020counterfactual}.
They use the learned reward function to change the reward of an actor-critic algorithm to help propagate signal over long horizons.
In particular, similar to our long-term setting, some works have specifically shown such state-associative architectures can perform better in delayed reward settings \citep{arjona2018rudder, hung2019optimizing, liu2019sequence, raposo2021synthetic}.
In contrast, we allow these properties to naturally emerge in a transformer architecture, without having to explicitly learn a reward function or a critic.

\subsection{Conditional language generation}

Various works have studied guided generation for images~\citep{karras2019stylegan} and language~\citep{ghazvininejad2017hafez, weng2021conditional}. Several works~\citep{ficler2017controlling, hu2017toward, rajani2019explain, yu2016sequence,ziegler2019fine, keskar2019ctrl} have explored training or fine-tuning of models for controllable text generation.
Class-conditional language models can also be used to learn disciminators to guide generation~\citep{dathathri2019plug, ghazvininejad2017hafez, holtzman2018learning,  krause2020gedi}.
However, these approaches mostly assume constant ``classes'', while in reinforcement learning the reward signal is time-varying.
Furthermore, it is more natural to prompt the model desired target return and continuously decrease it by the observed rewards over time, since the transformer model and environment jointly generate the trajectory.

\subsection{Attention and transformer models}
Transformers~\citep{vaswani2017attention} have been applied successfully to many tasks in natural language processing \citep{devlin2018bert,radford2018gpt} and computer vision \citep{carion2020detr,dosovitskiy2020vit}.
However, transformers are relatively unstudied in RL, mostly due to differing nature of the problem, such as higher variance in training.
\citet{zambaldi2018deep} showed that augmenting transformers with relational reasoning improve performance in combinatorial environments and \citet{ritter2020epn} showed iterative self-attention allowed for RL agents to better utilize episodic memories.
\citet{parisotto2020stabilizing} discussed design decisions for more stable training of transformers in the high-variance RL setting.
Unlike our work, these still use actor-critic algorithms for optimization, focusing on novelty in architecture.
Additionally, in imitation learning, some works have studied transformers as a replacement for LSTMs: \citet{dasari2020transformers} study one-shot imitation learning, and \citet{abramson2020imitating} combine language and image modalities for text-conditioned behavior generation.

%% file: sections/conclusion.tex
\section{Conclusion}

We proposed Decision Transformer, seeking to unify ideas in language/sequence modeling and reinforcement learning.
On standard offline RL benchmarks, we showed Decision Transformer can match or outperform strong algorithms designed explicitly for offline RL with minimal modifications from standard language modeling architectures.

We hope this work inspires more investigation into using large transformer models for RL. 
We used a simple supervised loss that was effective in our experiments, but applications to large-scale datasets could benefit from self-supervised pretraining tasks.
In addition, one could consider more sophisticated embeddings for returns, states, and actions -- for instance, conditioning on return distributions to model stochastic settings instead of deterministic returns.
Transformer models can also be used to model the state evolution of trajectory, potentially serving as an alternative to model-based RL, and we hope to explore this in future work.

For real-world applications, it is important to understand the types of errors transformers make in MDP settings and possible negative consequences, which are underexplored.
It will also be important to consider the datasets we train models on, which can potentially add destructive biases, particularly as we consider studying augmenting RL agents with more data which may come from questionable sources.
For instance, reward design by nefarious actors can potentially generate unintended behaviors as our model generates behaviors by conditioning on desired returns.

%% file: sections/acknowledgements.tex
\section*{Acknowledgements}

This research was supported by Berkeley Deep Drive, Open Philanthropy, and the National Science Foundation under NSF:NRI \#2024675. Part of this work was completed when Aravind Rajeswaran was a PhD student at the University of Washington, where he was supported by the J.P. Morgan PhD Fellowship in AI (2020-21). We also thank Luke Metz and Daniel Freeman for valuable feedback and discussions, as well as Justin Fu for assistance in setting up D4RL benchmarks, and Aviral Kumar for assistance with the CQL baselines and hyperparameters.

%% file: appendix/atari_hyperparameters.tex
\section{Experimental Details}

% Code for experiments can be found at: \texttt{\url{https://github.com/kzl/decision-transformer}}.

Code for experiments can be found in the supplementary material.

\subsection{Atari} \label{appendix:atari_hyperparameters}

We build our Decision Transformer implementation for Atari games off of minGPT (\url{https://github.com/karpathy/minGPT}), a publicly available re-implementation of GPT. We use most of the default hyperparameters from their character-level GPT example (\url{https://github.com/karpathy/minGPT/blob/master/play_char.ipynb}).
We reduce the batch size (except in Pong), block size, number of layers, attention heads, and embedding dimension for faster training.
For processing the observations, we use the DQN encoder from \citet{mnih2015human} with an additional linear layer to project to the embedding dimension.

% For return-to-go conditioning, we use either $1\times$ or $1.5\times$ CQL performance. We selected these values as our primary focus was comparison to CQL, but more possibilities exist for principled return-to-go conditioning.
For return-to-go conditioning, we use either $1\times$ or $5\times$ the maximum return in the dataset, but more possibilities exist for principled return-to-go conditioning.
In Atari experiments, we use Tanh instead of LayerNorm (as described in Section \ref{section:method}) after embedding each modality, but did this does not make a significant difference in performance.
The full list of hyperparameters can be found in Table \ref{tbl:atari_hyperparameters}.

% \begin{table}[ht]
% \caption{Hyperparameters of DT (and \%BC) for Atari experiments.}
% \vskip 0.15in
% \begin{center}
% \begin{small}
% \begin{tabular}{ll}
% \toprule
% \textbf{Hyperparameter} & \textbf{Value}  \\
% \midrule
% Number of layers & $6$  \\ 
% Number of attention heads    & $8$  \\
% Embedding dimension    & $128$  \\ 
% Batch size   & $512$ Pong \\
% & $128$ Breakout, Qbert, Seaquest\\ 
% Context length $K$ & $50$ Pong \\
% & $30$ Breakout, Qbert, Seaquest\\
% Return-to-go conditioning & $90$ Breakout ($\approx 1.5\times$ CQL performance) \\
% & $14000$ Qbert ($\approx 1\times$ CQL performance) \\
% & $20$ Pong ($\approx 1\times$ CQL performance) \\
% & $1150$ Seaquest ($\approx 1.5\times$ CQL performance) \\
% Nonlinearity & ReLU, encoder \\
% & GeLU, otherwise \\
% Encoder channels & $32, 64, 64$ \\
% Encoder filter sizes & $8 \times 8, 4 \times 4, 3 \times 3$ \\
% Encoder strides & $4, 2, 1$ \\
% Max epochs & $5$ \\
% Dropout & $0.1$ \\
% Learning rate & $6*10^{-4}$ \\
% Adam betas & $(0.9, 0.95)$ \\
% Grad norm clip & $1.0$ \\
% Weight decay & $0.1$ \\
% Learning rate decay & Linear warmup and cosine decay (see code for details) \\
% Warmup tokens & $512*20$ \\
% Final tokens & $2*500000*K$ \\

% \bottomrule
% \end{tabular}
% \end{small}
% \label{tbl:atari_hyperparameters}
% \end{center}
% % \vskip -0.1in
% \end{table} 

% not using CQL numbers for conditioning
\begin{table}[ht]
\caption{Hyperparameters of DT (and \%BC) for Atari experiments.}
\vskip 0.15in
\begin{center}
\begin{small}
\begin{tabular}{ll}
\toprule
\textbf{Hyperparameter} & \textbf{Value}  \\
\midrule
Number of layers & $6$  \\ 
Number of attention heads    & $8$  \\
Embedding dimension    & $128$  \\ 
Batch size   & $512$ Pong \\
& $128$ Breakout, Qbert, Seaquest\\ 
Context length $K$ & $50$ Pong \\
& $30$ Breakout, Qbert, Seaquest\\
Return-to-go conditioning & $90$ Breakout ($\approx 1\times$ max in dataset) \\
& $2500$ Qbert ($\approx 5\times$ max in dataset) \\
& $20$ Pong ($\approx 1\times$ max in dataset) \\
& $1450$ Seaquest ($\approx 5\times$ max in dataset) \\
Nonlinearity & ReLU, encoder \\
& GeLU, otherwise \\
Encoder channels & $32, 64, 64$ \\
Encoder filter sizes & $8 \times 8, 4 \times 4, 3 \times 3$ \\
Encoder strides & $4, 2, 1$ \\
Max epochs & $5$ \\
Dropout & $0.1$ \\
Learning rate & $6*10^{-4}$ \\
Adam betas & $(0.9, 0.95)$ \\
Grad norm clip & $1.0$ \\
Weight decay & $0.1$ \\
Learning rate decay & Linear warmup and cosine decay (see code for details) \\
Warmup tokens & $512*20$ \\
Final tokens & $2*500000*K$ \\

\bottomrule
\end{tabular}
\end{small}
\label{tbl:atari_hyperparameters}
\end{center}
% \vskip -0.1in
\end{table} 

\subsection{OpenAI Gym}

\subsubsection{Decision Transformer}

Our code is based on the Huggingface Transformers library \citep{wolf-etal-2020-transformers}.
Our hyperparameters on all OpenAI Gym tasks are shown below in Table \ref{tbl:gym_hyperparameters}.
Heuristically, we find using larger models helps to model the distribution of returns, compared to standard RL model sizes (which learn one policy).
For reacher we use a smaller context length than the other environments, which we find to be helpful as the environment is goal-conditioned and the episodes are shorter.
We choose return targets based on expert performance for each environment, except for HalfCheetah where we find 50\% performance to be better due to the datasets containing lower relative returns to the other environments.
Models were trained for $10^5$ gradient steps using the AdamW optimizer \citep{loshchilov2017adamw} following PyTorch defaults.

\begin{table}[ht]
\caption{Hyperparameters of Decision Transformer for OpenAI Gym experiments.}
\vskip 0.15in
\begin{center}
\begin{small}
\begin{tabular}{ll}
\toprule
\textbf{Hyperparameter} & \textbf{Value}  \\
\midrule
Number of layers & $3$  \\ 
Number of attention heads    & $1$  \\
Embedding dimension    & $128$  \\
Nonlinearity function & ReLU \\
Batch size   & $64$ \\
Context length $K$ & $20$ HalfCheetah, Hopper, Walker \\
               & $5$ Reacher \\
Return-to-go conditioning   & $6000$ HalfCheetah \\
                            & $3600$ Hopper \\
                            & $5000$ Walker \\
                            & $50$ Reacher \\
Dropout & $0.1$ \\
Learning rate & $10^{-4}$ \\
Grad norm clip & $0.25$ \\
Weight decay & $10^{-4}$ \\
Learning rate decay & Linear warmup for first $10^5$ training steps \\
\bottomrule
\end{tabular}
\end{small}
\label{tbl:gym_hyperparameters}
\end{center}
% \vskip -0.1in
\end{table} 

\subsubsection{Behavior Cloning}

As briefly mentioned in Section \ref{sec:gym_results}, we found previously reported behavior cloning baselines to be weak, and so run them ourselves using a similar setup as Decision Transformer.
We tried using a transformer architecture, but found using an MLP (as in previous work) to be stronger.
We train for $2.5 \times 10^4$ gradient steps; training more did not improve performance.
Other hyperparameters are shown in Table \ref{tbl:gym_bc_hyperparameters}.
The percentile behavior cloning experiments use the same hyperparameters.

\begin{table}[ht]
\caption{Hyperparameters of Behavior Cloning for OpenAI Gym experiments.}
\vskip 0.15in
\begin{center}
\begin{small}
\begin{tabular}{ll}
\toprule
\textbf{Hyperparameter} & \textbf{Value}  \\
\midrule
Number of layers & $3$ \\
Embedding dimension & $256$  \\ 
Nonlinearity function & ReLU \\
Batch size   & $64$ \\
Dropout & $0.1$ \\
Learning rate & $10^{-4}$ \\
Weight decay & $10^{-4}$ \\
Learning rate decay & Linear warmup for first $10^5$ training steps \\
\bottomrule
\end{tabular}
\end{small}
\label{tbl:gym_bc_hyperparameters}
\end{center}
% \vskip -0.1in
\end{table} 

\subsection{Graph Shortest Path}

We give details of the illustrative example discussed in the introduction. The task is to find the shortest path on a fixed directed graph, which can be formulated as an MDP where reward is $0$ when the agent is at the goal node and $-1$ otherwise. The observation is the integer index of the graph node the agent is in. The action is the integer index of the graph node to move to next. The transition dynamics transport the agent to the action's node index if there is an edge in the graph, while the agent remains at the past node otherwise. The returns-to-go in this problem correspond to negative path lengths and maximizing them corresponds to generating shortest paths. 

In this environment, we use the GPT model as described in Section \ref{section:method} to generate both actions and return-to-go tokens. This makes it possible for the model it generate its own (realizable) returns-to-go $\hat{R}$. Since we require a return prompt to generate actions and we do assume knowledge of the optimal path length upfront, we use a simple prior over returns that favors shorter paths: $P_\text{prior}(\hat{R}=k) \propto T+1-k$, where $T$ is the maximum trajectory length. Then, it is combined with the return probabilities generated by the GPT model: $P(\hat{R}_t | s_{0:t}, a_{0:t-1}, \hat{R}_{0:t-1}) = P_\text{GPT}(\hat{R}_t | s_{0:t}, a_{0:t-1}, \hat{R}_{0:t-1}) \times P_\text{prior}(\hat{R}_t)^{10}$. Note that the prior and return-to-go predictions are entirely computable by the model, and thus avoids the need for any external or oracle information like the optimal path length. Adjustment of generation by a prior has also been used for similar purposes in controllable text generation in prior work~\citep{krause2020gedi}. 

\begin{figure*}
\centering
\includegraphics[width=0.6\textwidth]{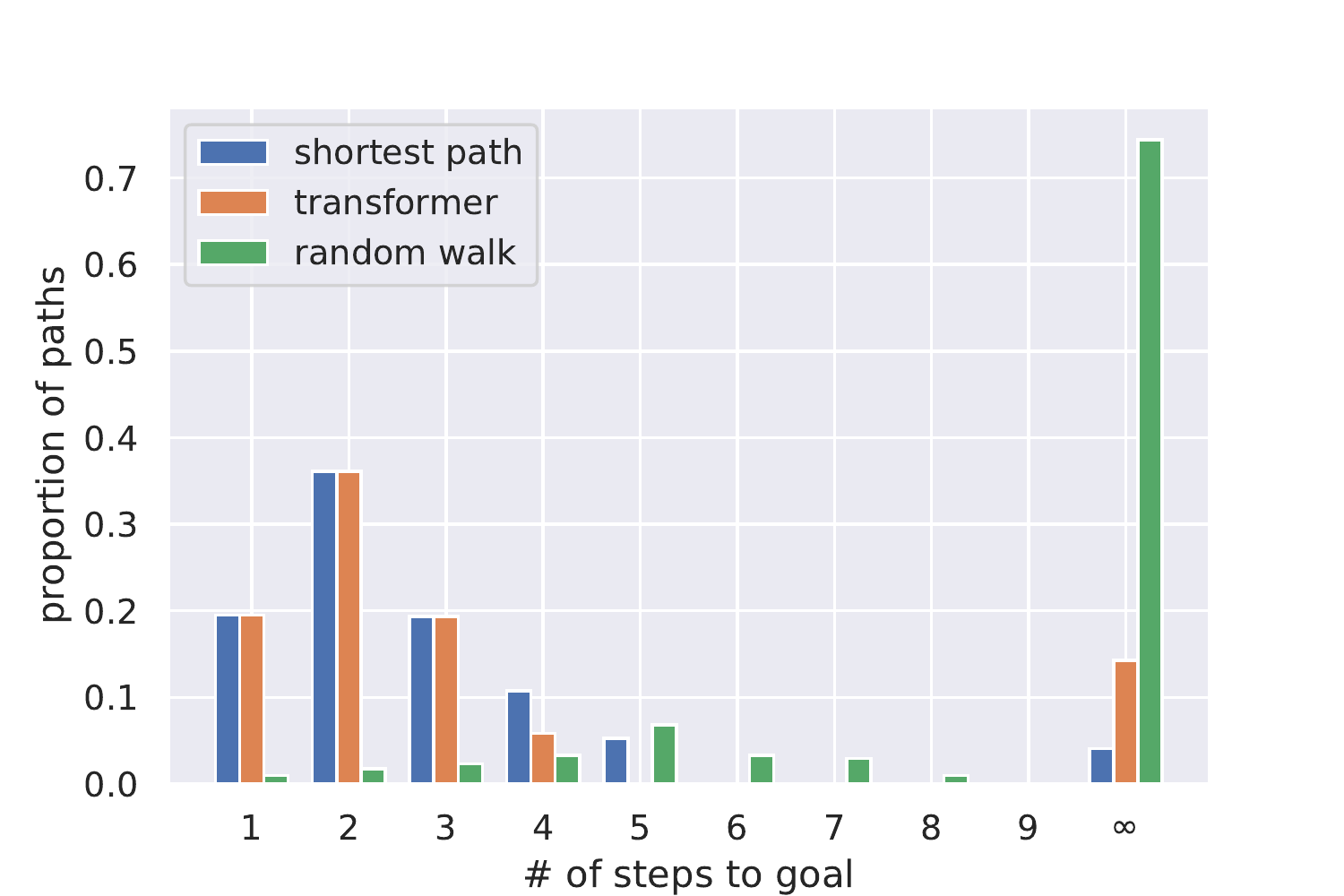}
\caption{Histogram of steps to reach the goal node for random walks on the graph, shortest possible paths to the goal, and attempted shortest paths generated by the transformer model. $\infty$ indicates the goal was not reached during the trajectory.}
\label{fig:graph_length}
\end{figure*}

We train on a dataset of $1,000$ graph random walk trajectories of $T=10$ steps each with a random graph of $20$ nodes and edge sparsity coefficient of $0.1$. We report the results in Figure \ref{fig:graph_length}, where we find that transformer model is able to significantly improve upon the number of steps required to reach the goal, closely matching performance of optimal paths.

There are two reasons for the favorable performance on this task. In one case, the training dataset of random walk trajectories may contain a segment that directly corresponds to the desired shortest path, in which case it will be generated by the model. In the second case, generated paths are entirely original and are not subsets of trajectories in the training dataset - they are generated from stitching sub-optimal segments. We find this case accounts for $15.8\%$ of generated paths in the experiment.

While this is a simple example and uses a prior on generation that we do not use in other experiments for simplicity, it illustrates how hindsight return information can be used with generation priors to avoid the need for explicit dynamic programming.

%% file: appendix/atari_raw.tex
\section{Atari Task Scores} \label{appendix:atari_raw}
Table \ref{tbl:atari_baselines} shows the normalized scores used for normalization used in \citet{hafner2020mastering}. Tables \ref{tbl:atari_main_raw} and \ref{tbl:atari_percentile_bc_raw} show the raw scores corresponding to Tables \ref{tbl:atari_main} and \ref{tbl:atari_percentile_bc}, respectively. For \%BC scores, we use the same hyperparameters as Decision Transformer for fair comparison.  For REM and QR-DQN, there is a slight discrepancy between \citet{agarwal2020optimistic} and \citet{kumar2020conservative}; we report raw data provided to us by REM authors.

\begin{table*}[h]
\centering
\small
\begin{tabular}{lrr}
\toprule
\multicolumn{1}{c}{\bf Game} & \multicolumn{1}{c}{\bf Random} & \multicolumn{1}{c}{\bf Gamer} \\
  \midrule
Breakout & $2$ & $30$ \\
Qbert & $164$ & $13455$ \\ 
Pong & $-21$ & $15$ \\
Seaquest & $68$ & $42055$ \\
 \bottomrule
 \end{tabular}
 \caption{Atari baseline scores used for normalization.}
 \label{tbl:atari_baselines}
\end{table*}

% \begin{table*}[h]
% \centering
% \small
% \begin{tabular}{lrrrrr}
% \toprule
% \multicolumn{1}{c}{\bf Game} & \multicolumn{1}{c}{\bf DT (Ours)} & \multicolumn{1}{c}{\bf CQL} & \multicolumn{1}{c}{\bf QR-DQN} & \multicolumn{1}{c}{\bf REM} & \multicolumn{1}{c}{\bf BC} \\
% \midrule
% Breakout  & $\bf{76.9} \pm 27.3$ & $61.1$ & $7.9$ & $11.0$& $40.9 \pm 17.3$ \\
% Qbert     & $3498.3 \pm 2402.1$ & $\bf{14012.0}$ & $383.6$ & $343.4$ & $2464.1 \pm 1948.2$ \\
% Pong      & $17.1 \pm 2.9$ & $\bf{19.3}$ & $-13.8$ & $-6.9$ & $9.7 \pm 7.2$ \\
% Seaquest  & $\bf{1063.3} \pm 274.8$ & $779.4$ & $672.9$  & $499.8$   & $968.6 \pm 133.8$ \\
% \bottomrule
% \end{tabular}
% \caption{
% Raw scores for the 1\% DQN-replay Atari dataset.
% We report the mean and variance across 3 seeds.
% Best mean scores are highlighted in bold.
% Decision Transformer (DT) performs comparably to CQL on 3 out of 4 games, and outperforms other baselines.}
% \label{tbl:atari_main_raw}
% \end{table*}

% updated QR-DQN / REM numbers as per Rishabh Agarwal's email (results from CQL paper were incorrect)
% also using RTG conditioning not based on CQL numbers
\begin{table*}[h]
\centering
\small
\begin{tabular}{lrrrrr}
\toprule
\multicolumn{1}{c}{\bf Game} & \multicolumn{1}{c}{\bf DT (Ours)} & \multicolumn{1}{c}{\bf CQL} & \multicolumn{1}{c}{\bf QR-DQN} & \multicolumn{1}{c}{\bf REM} & \multicolumn{1}{c}{\bf BC} \\
\midrule
Breakout  & $\bf{76.9} \pm 27.3$ & $61.1$ & $6.8$ & $4.5$& $40.9 \pm 17.3$ \\
Qbert     & $2215.8 \pm 1523.7$ & $\bf{14012.0}$ & $156.0$ & $160.1$ & $2464.1 \pm 1948.2$ \\
Pong      & $17.1 \pm 2.9$ & $\bf{19.3}$ & $-14.5$ & $-20.8$ & $9.7 \pm 7.2$ \\
Seaquest  & $\bf{1129.3} \pm 189.0$ & $779.4$ & $250.1$  & $370.5$   & $968.6 \pm 133.8$ \\
\bottomrule
\end{tabular}
\caption{
Raw scores for the 1\% DQN-replay Atari dataset.
We report the mean and variance across 3 seeds.
Best mean scores are highlighted in bold.
Decision Transformer performs comparably to CQL on 3 out of 4 games, and usually outperforms other baselines.}
\label{tbl:atari_main_raw}
\end{table*}

\begin{table*}[h]
\centering
\small
\begin{tabular}{lrrrrrr}
\toprule
\multicolumn{1}{c}{\bf Game} & \multicolumn{1}{c}{\bf DT (Ours)} & \multicolumn{1}{c}{\bf 10\%BC} & \multicolumn{1}{c}{\bf 25\%BC} & \multicolumn{1}{c}{\bf 40\%BC} & \multicolumn{1}{c}{\bf 100\%BC} \\
  \midrule
Breakout  & $\bf{76.9} \pm 27.3$ & $10.0 \pm 2.3$ & $22.6 \pm 1.8$ & $32.3 \pm 18.9$ &  $40.9 \pm 17.3$ \\
Qbert    & $2215.8 \pm 1523.7$ & $1045 \pm 232.0$ & $2302.5 \pm 1844.1$ & $1674.1 \pm 776.0$ & $\bf{2464.1} \pm 1948.2$ \\
Pong      & $\bf{17.1} \pm 2.9$ & $-20.3 \pm 0.1$ & $-16.2 \pm 1.0$ & $5.2 \pm 4.8$ & $9.7 \pm 7.2$ \\
Seaquest  & $\bf{1129.3} \pm 189.0$ & $521.3 \pm 103.0$ & $549.3 \pm 96.2$ & $758 \pm 169.1$ & $968.6 \pm 133.8$ \\
\bottomrule
\end{tabular}
\caption{\%BC scores for Atari. We report the mean and variance across 3 seeds.
Decision Transformer usually outperforms \%BC.}
\label{tbl:atari_percentile_bc_raw}
\end{table*}